\documentclass[letterpaper]{article} 
\usepackage{aaai23}  
\usepackage{times}  
\usepackage{helvet}  
\usepackage{courier}  
\usepackage[hyphens]{url}  
\usepackage{graphicx} 
\usepackage{natbib}  
\usepackage{caption} 
\usepackage{algorithm}
\usepackage{algorithmic}
\usepackage{relsize}
\usepackage{amsfonts,amssymb}
\usepackage[fleqn]{amsmath}
\usepackage{tikz}
\usepackage{subcaption}
\usepackage{booktabs}
\usepackage{multirow}
\usetikzlibrary{shapes.geometric}

\usepackage{newfloat}
\usepackage{listings}
\DeclareCaptionStyle{ruled}{labelfont=normalfont,labelsep=colon,strut=off} 
\floatstyle{ruled}
\newfloat{listing}{tb}{lst}{}
\floatname{listing}{Listing}
 \nocopyright 

\title{Learning Lagrangian Multipliers for the Travelling Salesman Problem}
\author{
    Augustin Parjadis,\textsuperscript{\rm 1}
    Quentin Cappart,\textsuperscript{\rm 1}
    Bistra Dilkina,\textsuperscript{\rm 2}\\
    Aaron Ferber,\textsuperscript{\rm 2}
    Louis-Martin Rousseau\textsuperscript{\rm 1}}

\affiliations {
    \textsuperscript{\rm 1} Polytechnique Montréal, Montreal, Canada\\
    \textsuperscript{\rm 2} Center for Artificial Intelligence in Society, University of Southern California, USA\\
    \{augustin.parjadis-de-lariviere, quentin.cappart, louis-martin.rousseau\}@polymtl.ca\\
    \{aferber, dilkina\}@usc.edu
}

\begin{document}

\maketitle

\begin{abstract}
Lagrangian relaxation is a versatile mathematical technique employed to relax constraints in an optimization problem, enabling the generation of dual bounds to prove the optimality of feasible solutions and the design of efficient propagators in constraint programming (such as the weighted circuit constraint).
However, the conventional process of deriving Lagrangian multipliers (e.g., using subgradient methods) is often computationally intensive, limiting its practicality for large-scale or time-sensitive problems.
To address this challenge, we propose an innovative unsupervised learning approach that harnesses the capabilities of graph neural networks to exploit the problem structure, aiming to generate accurate Lagrangian multipliers efficiently.
We apply this technique to the well-known Held-Karp Lagrangian relaxation for the travelling salesman problem. The core idea is to predict accurate Lagrangian multipliers and to employ them as a warm start for generating Held-Karp relaxation bounds. These bounds are subsequently utilized to enhance the filtering process carried out by branch-and-bound algorithms. In contrast to much of the existing literature, which primarily focuses on finding feasible solutions, our approach operates on the dual side, demonstrating that learning can also accelerate the proof of optimality.
We conduct experiments across various distributions of the metric travelling salesman problem, considering instances with up to 200 cities. The results illustrate that our approach can improve the filtering level of the weighted circuit global constraint, reduce the optimality gap by a factor two for unsolved instances up to a timeout,
and reduce the execution time for solved instances by 10\%.
\end{abstract}

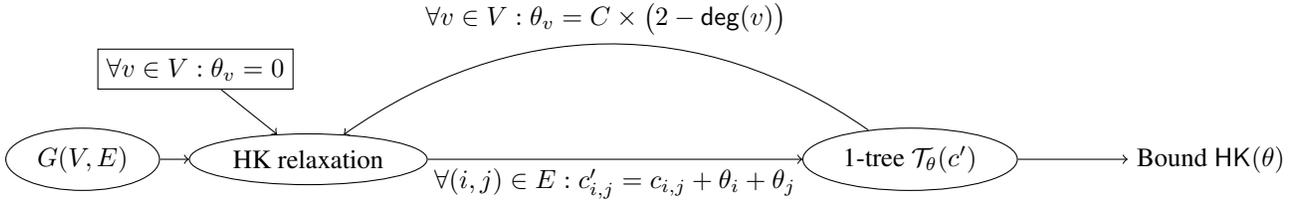
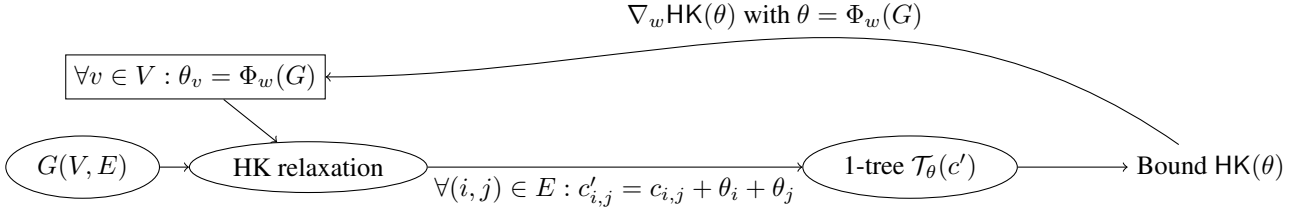
\begin{figure*}[ht!]
  \centering

\begin{subfigure}[b]{\textwidth}
\begin{tikzpicture}
  \node[ellipse, draw] (e1) at (0,0) {$G(V,E)$};
  \node[draw] (et) at (1.5,1.2) {$\forall v\in V: \theta_v = 0$};
  \node[ellipse, draw] (e2) at (3,0) {HK relaxation};
  \node[ellipse, draw] (e3) at (11,0) {1-tree $\mathcal{T}_\theta(c^\prime)$};
  \node (eb) at (15,0) {Bound  $\mathsf{HK}(\theta)$};
  
  \draw[->] (e1) -- (e2);
  \draw[->] (et) -- (e2);
  \draw[->] (e2) to node [midway, below] {$\forall (i,j) \in E: c^\prime_{i,j} = c_{i,j} + \theta_i + \theta_j$} (e3);
  \draw[->] (e3) -- (eb);
  \draw[->] (e3) to [out=145,in=35] node [midway, above] {$\forall v\in V:  \theta_v = C \times \big(2 - \mathsf{deg}(v) \big)$} (e2);
\end{tikzpicture}
  \caption{Approach of \citet{heldkarp70,heldkarp71} - Iterative process for improving $\theta$ multipliers.} \label{fig:HKrelaxDiag}
\end{subfigure}

\begin{subfigure}[b]{\textwidth}
\begin{tikzpicture}
  \node[ellipse, draw] (e1) at (0,0) {$G(V,E)$};
  \node[draw] (et) at (1.5,1.2) {$\forall v\in V: \theta_v= \Phi_w(G)$};
  \node[ellipse, draw] (e2) at (3,0) {HK relaxation};
  \node[ellipse, draw] (e3) at (11,0) {1-tree $\mathcal{T}_\theta(c^\prime)$};
  \node (eb) at (15,0) {Bound $\mathsf{HK}(\theta)$};
  
  \draw[->] (e1) -- (e2);
  \draw[->] (et) -- (e2);
  \draw[->] (e2) to node [midway, below] {$\forall (i,j) \in E: c^\prime_{i,j} = c_{i,j} + \theta_i + \theta_j$} (e3);
  \draw[->] (e3) -- (eb);
  \draw[->] (eb) to [out=145,in=0] node [midway, above] {$\nabla_w \mathsf{HK}( \theta) ~ \text{with}  ~ \theta = \Phi_w(G)$ } (et);
\end{tikzpicture}
  \caption{Our contribution - Unsupervised learning approach to obtain $\theta$ multipliers through backpropagation.} \label{fig:HKGNNDiag}
\end{subfigure}
\caption{Illustration of both procedures to obtain dual bounds for the TSP.}
\end{figure*}

\section{Introduction}

The \textit{travelling salesman problem} (TSP), although simple, has been the subject of extensive research and has broad practical applications. 
Due to its NP-hard nature, numerous approaches have been proposed to solve it efficiently, ranging from exact to heuristic methods~\cite{lawler1986tspsurvey}. 
Exact solvers not only need to identify the optimal solution but also to prove that it is optimal, often via a dual bound. \citet{heldkarp70} proposed a relaxation that provides strong dual bounds in practice. For instance, these bounds are used in Concorde, the state-of-the-art TSP solver~\cite{Applegate06} or in the design
of global constraints in constraint programming~\cite{benchimol2010improving,benchimol2012improved}. 
An associated branch-and-bound algorithm using this relaxation was subsequently proposed by \citet{heldkarp71}, 
which enabled the optimality proof for several open benchmark instances at the time of its publication. Briefly, this algorithm leverages a combinatorial structure, referred to as \textit{minimum 1-tree}, that can serve as a valid relaxation for the TSP and obtain dual bounds.
However, this algorithm is based on a few heuristic design choices which have an important impact on the tightness of the relaxation. One is the procedure to generate the bounds from Lagrangian multipliers (explained in the next section), which can be assimilated as a hill-climbing algorithm. 
Starting from initial bounds, the algorithm refines the bound iteratively with local perturbations until convergence. There are two drawbacks to this process. First, it requires several potentially costly iterations to get accurate bounds, and second, it only converges to local minima.
Our research hypothesis is that this procedure can be improved thanks to a learning-based approach. The idea
is to train a model in an unsupervised fashion with similar TSP instances and to use it to predict  Lagrangian multipliers that can be used to obtain a valid dual bound instead of computing it iteratively.

Machine learning has helped guide heuristic components in branch-and-bound~\cite{Khalil_Dilkina_2016,lodi2017learning,gasse2019exact,Yilmaz_2021}, constraint programming~\cite{cappart2021combining,chalumeau2021seapearl}, SAT solving~\cite{selsam2018learning,selsam2019guiding}, local search~\cite{d2020learning,xin2021neurolkh}, and non linear optimization~\cite{ferber2023surco}. We refer to the survey of \citet{bengio2020machine} for an extended 
literature review on this topic. Most of such works operate on the branching decisions (e.g., selecting the next variable to branch on) or on the primal side. However, learning to improve the quality of  relaxations by means of better dual bounds has been much less considered in the literature. To our knowledge, this has only been addressed
for the restricted use case of solvers
based on decision diagrams~\cite{cappart2019improving,cappart2022improving,parjadis2021improving}.

Additionally, recent work in \textit{decision-focused learning} (DFL) has approached settings where the problem formulation is not fully specified at the time of decision-making. Thus, these approaches train gradient-based deep learning models to predict the missing components, with a key component being to determine how to train the deep learning model to improve the downstream decision quality. As training for deep networks is done using gradient descent, the difficulty lies in deriving methods for differentiating the output of the optimization model with respect to its predicted inputs. Our proposed approach seeks to predict the parameters of the Held Karp relaxation such that the resulting relaxed solution provides a dual bound as tight as possible. This is achieved by deriving gradients for the relaxation to learn parameters that directly optimize the related bound. Differentiation has been successfully deployed for quadratic programs \cite{amos17}, linear programs \cite{wilder2019melding, elmachtoub2020smart, mandi2020interior, mandi2020smart}, mixed integer linear programs \cite{ferber2019mipaal}, MAXSAT \cite{wang2019satnet}, and blackbox discrete optimization \cite{poganvcic2019diffbb, niepert2021imle}, among others discussed in these surveys \cite{sadana2023diffoptsurvey,ijcai2021p610}. However, this approach is the first to consider using differentiable optimization to improve exact solver efficiency.

Coming back to the TSP, the design of learning-based 
solving approaches has also sparked a great interest in the research community~\cite{bello2016neural,deudon2018learning,kool2018attention}.
In an industrial context, this methodology is relevant for
practitioners who are solving similar problem instances every day and want to leverage historical decisions, e.g. in last-mile package delivery~\cite{merchan20222021}. \textit{Graph neural networks} (GNN) is a neural architecture~\cite{scarselli2008graph,kipf2016semi} widely considered for the TSP~\cite{joshi2022learning}. More generally, GNNs also play a crucial role in the success of applying deep learning to combinatorial optimization \cite{khalil2017learning,cappart2021combinatorial}. They allow for the extraction of rich hidden representations by successively aggregating the weights of neighboring nodes in a graph, on which many combinatorial problems are defined. 

Based on this context, the contribution of the paper is an approach 
based on unsupervised learning and graph neural networks 
to generate appropriate Lagrangian multipliers for the TSP, 
which are then used to improve the Held-Karp relaxation.
Additionally, we integrate this mechanism inside a branch-and-bound algorithm with domain filtering and constraint propagation \cite{benchimol2012improved} to improve exact TSP solving.
Experiments are carried out on three distributions of metric TSPs and the results show that
our approach can improve the filtering level of the weighted circuit global constraint, reduce the optimality gap by a factor of two for unsolved instances up to a timeout,
and reduce the execution time for solved instances by 10\%.

The following section briefly overviews the Held-Karp relaxation principle for the TSP. Building upon this, we next describe the proposed learning approach for generating bounds through unsupervised learning on the Lagrangian multipliers of the Held-Karp relaxation. Finally, we discuss the training and integration of dual-bound generation within a branch-and-bound algorithm to evaluate their impact.

\begin{figure*}[ht!]
  \centering

\begin{subfigure}[b]{0.3\textwidth}
    \begin{tikzpicture}[scale=1.9]
    \foreach \i in {1,...,5}
      \node[circle, draw] (G1-\i) at (72*\i:1) {\i};
    
    \draw[dashed] (G1-1) -- node[right] {20} (G1-5);
    \draw[dashed] (G1-1) -- node[right] {22} (G1-4);
    \draw[dashed] (G1-5) -- node[below right] {15} (G1-4);
    \draw[dashed] (G1-5) -- node[below] {40} (G1-3);
    \draw[dashed] (G1-4) -- node[below] {14} (G1-3);
    \draw[dashed] (G1-1) -- node[above] {10} (G1-2);
    \draw[dashed] (G1-1) -- node[above left] {16} (G1-3);
    \draw[dashed] (G1-2) -- node[above] {7} (G1-5);
    \draw[dashed] (G1-2) -- node[below left] {12} (G1-4);
    \draw[dashed] (G1-2) -- node[left] {5} (G1-3);

\end{tikzpicture}
\caption{Initial TSP instance. \\ The optimal cost is 62.}
\label{sub:example-1}
\end{subfigure}
\hfill
\begin{subfigure}[b]{0.3\textwidth}
\begin{tikzpicture}[scale=1.9]
    \foreach \i in {1,...,5}
      \node[circle, draw] (G1-\i) at (72*\i:1) {\i};
      
    \draw[dashed] (G1-1) -- node[right] {20} (G1-5);
    \draw[dashed] (G1-1) -- node[right] {22} (G1-4);
    \draw[dashed] (G1-5) -- node[below right] {15} (G1-4);
    \draw[dashed] (G1-5) -- node[below] {40} (G1-3);
    \draw[dashed] (G1-4) -- node[below] {14} (G1-3);
    \draw[blue] (G1-1) -- node[above] {10} (G1-2);
    \draw[blue]  (G1-1) -- node[above left] {16} (G1-3);
    \draw[blue]  (G1-2) -- node[above] {7} (G1-5);
    \draw[blue]  (G1-2) -- node[below left] {12} (G1-4);
    \draw[blue]  (G1-2) -- node[left] {5} (G1-3);
    \end{tikzpicture}
\caption{Minimum 1-tree under the initial costs. \\ Lower bound obtained is 50 (gap of $20\%$)}
\label{sub:example-2}
\end{subfigure}
\hfill
\begin{subfigure}[b]{0.3\textwidth}
    \begin{tikzpicture}[scale=1.9]
    \foreach \i in {1,...,5}
      \node[circle, draw] (G2-\i) at (72*\i:1) {\i};
      
    
    \draw (G2-1)[dashed] -- node[right] {18} (G2-5);
    \draw (G2-1)[dashed] -- node[right] {20} (G2-4);
    \draw[blue] (G2-5) -- node[below right] {11} (G2-4);
    \draw[dashed] (G2-5) -- node[below] {38} (G2-3);
    \draw[dashed] (G2-4) -- node[below] {12} (G2-3);
    \draw[blue] (G2-1) -- node[above] {14} (G2-2);
    \draw[blue] (G2-1) -- node[above left] {16} (G2-3);
    \draw[blue] (G2-2) -- node[above] {9} (G2-5);
    \draw[dashed] (G2-2) -- node[below left] {14} (G2-4);
    \draw[blue] (G2-2) -- node[left] {9} (G2-3);

      \node[above=10,font=\bfseries] at (G2-2.center) {$\theta_2:4$};
      \node[above=10,font=\bfseries] at (G2-1.center) {$\theta_1:0$};
      \node[below=10,font=\bfseries] at (G2-3.center) {$\theta_3:0$};
      \node[right=10,font=\bfseries] at (G2-4.center) {$\theta_4:-2$};
      \node[below=10,xshift=1em,font=\bfseries] at (G2-5.center) {$\theta_5:-2$};
\end{tikzpicture}
\caption{Minimum 1-tree with modified costs. \\ Lower bound obtained is  59 (gap of $5\%$)}
\label{sub:example-3}
\end{subfigure}
  \hfill
  \caption{Illustration of a single iteration of Held-Karp relaxation for an arbitrary TSP instance.} \label{fig:HKrelax}
\end{figure*}
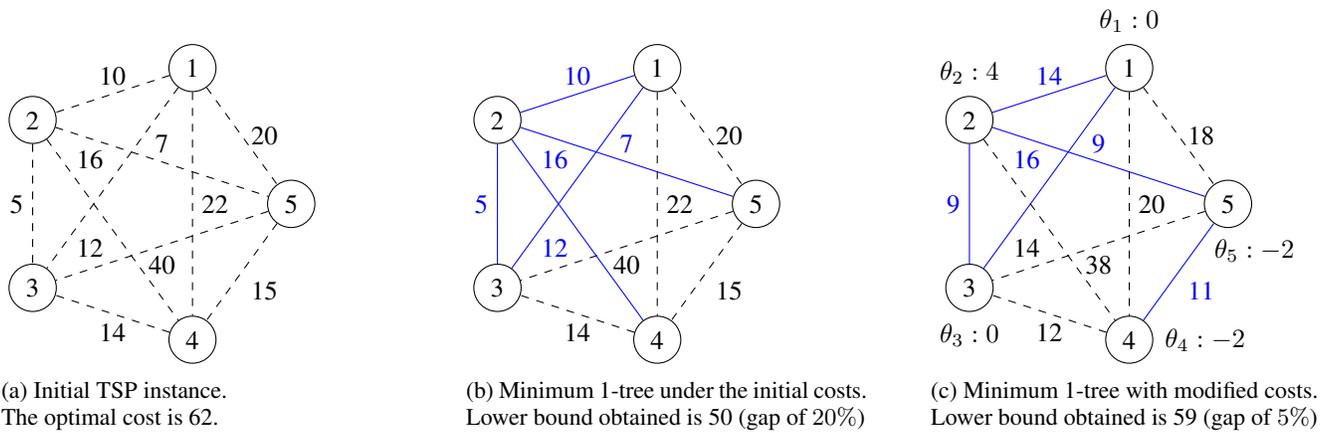

\section{Held-Karp Lagrangian Relaxation} \label{sec:relaxHK}

Finding optimal solutions for large TSP instances requires sophisticated approaches due to the combinatorial explosion of the solution space. With branch-and-bound, optimization bounds are employed to prune the search tree and accelerate the search, allowing solvers to prove optimality without exploring the entire tree. To achieve this, the \textit{Held-Karp relaxation} \cite{heldkarp70} offers a robust dual bound based on a variant of minimum spanning trees.

Let $G = (V, E)$ be a complete graph with a cost  attached to each edge. A \textit{minimum 1-tree} is a minimum spanning tree of $G\backslash\{1\}$ to which we add the node $1$ along with the two cheapest edges connecting it to the tree.
We note that the choice of node $1$ is arbitrary, 
depending on the labeling of $V$. 
A minimum 1-tree can be obtained by solving the integer program presented in Equations~\eqref{eq:l1} to \eqref{eq:l5}.
Constraints~\eqref{eq:l2} and~\eqref{eq:l3} define the 1-tree structure and
Constraint~\eqref{eq:l4} enforces the elimination of sub-tours. 
This problem involves finding a minimum spanning tree that can be solved in $\mathcal{O}(E\log V)$ by Kruskal's algorithm. Here, $\delta(v)$ denotes the edges containing vertex $v \in V$. We use $c_e \in \mathbb{R}$  to represent the cost of an edge $e \in E$, and  $x_e \in \{0,1\}$ 
is the decision variable indicating whether edge $e$ is included in the 1-tree.
\begin{align}
\min ~  & \mathlarger{\sum_{e \in E}} c_e x_e  & \label{eq:l1}\\
\text{s. t.} ~ & \mathlarger{\sum_{e \in \delta(1)}} x_e = 2 & \label{eq:l2}\\
& \mathlarger{\sum_{e \in E}} ~ x_e = |V| & \label{eq:l3}\\
& \mathlarger{\sum_{\substack{i,j \in S \\ i<j}}} x_{i,j} \leq |S|-1 & \forall S \subset V \backslash \{1\}  \land  
|S| \geq 3  \label{eq:l4}\\
& x_e \in \{0,1\} & \forall e \in E  \label{eq:l5}
\end{align}
Let us note that every tour in $G$ is a 1-tree, 
and if a minimum  1-tree is a tour,
it is an optimal solution to the TSP. 
Therefore any minimum 1-tree is a valid relaxation for the TSP,
which is an interesting property to leverage.
However, a solution of this integer program is not ensured to be a tour. To
do so, a new set of constraints must be enforced.
\begin{equation}
\label{eq:l55}
  \sum_{e \in \delta(v)} x_e = 2  \hspace{3cm} \forall v \in V \backslash \{1\}  
\end{equation}
These constraints force each node to have only two edges, an incoming and an outgoing one, and turn the problem in 
finding a minimum-cost Hamiltonian cycle, which is NP-hard.
To obtain a valid 1-tree relaxation efficiently, one can then move
these constraints (one for each node) into the Objective~\eqref{eq:l1} and penalize their violations with associated Lagrangian multipliers $\theta_v \in \mathbb{R}$ for each $v \in V \backslash \{1\}$. The updated objective function is as follows.
\begin{equation}
\label{eq:l6} 
    \min \sum_{e \in E} c_e x_e - \sum_{v \in V \backslash \{1\}} \theta_v \Bigg(2 - \sum_{e \in \delta(v)} x_e\Bigg) 
\end{equation}
Intuitively, each node having a degree other than two will be penalized.
An optimal 1-tree relaxation can be found by optimizing over the $\theta_v$ variables.
To do so, an iterative approach has been proposed by \citet{heldkarp70,heldkarp71}. The idea
is to adjust the Lagrangian multipliers $\theta$ step-by-step to build a sequence of 1-trees which provides increasingly better bounds. An overview of the process is proposed in Figure~\ref{fig:HKrelaxDiag}.
First, an initial minimum 1-tree is computed by finding a minimum spanning tree on $G \backslash \{1\}$ and adding the two cheapest edges incident to node $1$.
If the optimal 1-tree is a tour, it corresponds to the optimal TSP solution.
Otherwise, some constraints are penalized as at least one node has a degree greater than 2.
The main idea of \citet{heldkarp70,heldkarp71} is to penalize such nodes by modifying the cost $c_{i,j}$ of edges $(i,j)\in E$,
based on the values of $\theta_i$ and $\theta_j$ (i.e., the multipliers of adjacent nodes).
Let $c^\prime_{i,j} \in \mathbb{R}$ be the modified costs. They are computed as follows.
\begin{equation}
\label{eq:l7} 
c^\prime_{i,j} = c_{i,j} + \theta_i + \theta_j \hspace{2.5cm} \forall (i,j) \in E
\end{equation}
A theoretical property proved by \citet{heldkarp70,heldkarp71} is that the optimal TSP tour is invariant 
under this perturbation, whereas the optimal 1-tree is not. This gives room to improve the solution by 
finding better multipliers.
Equation~\eqref{eq:penalty} proposes a standard choice to compute the multiplier, where $C \in \mathbb{R}$ is an arbitrary constant and $\mathsf{deg}(v)$ denotes the degree of node $v \in V$ in the current 1-tree.
\begin{equation}
\label{eq:penalty} 
\theta_v = C \times \big(2 - \mathsf{deg}(v) \big) \hspace{2.5cm} \forall v \in V
\end{equation}
Finally, a new minimum 1-tree is computed from the graph with the updated costs $c^\prime_{i,j}$.
We note this 1-tree as $\mathcal{T}_\theta(c^\prime)$ where $c^\prime = \{c_1,\dots, c_{|E|}\}$ is the set of all modified costs, and $\theta = \{\theta_1,\dots,\theta_{|V|} \}$ is the set of all multipliers.
We also use the notation $\mathsf{cost}\big(\mathcal{T}_\theta(c^\prime)\big)$ to refer to the total cost of the 1-tree. 
This process is reiterated, and a new 1-tree $\mathcal{T}_\theta(c^\prime)$ is obtained until no improvement is obtained (i.e., when a local minimum is reached).
The cost of the  optimal 1-tree gives a lower bound on the objective value as follows.
\begin{equation}
\label{eq:bound} 
\mathsf{HK}(\theta) = \mathsf{cost}\big(\mathcal{T}_\theta(c^\prime)\big) - 2\sum_{i = 1}^{|V|} \theta_i
\end{equation}
This bound,  $\mathsf{HK}(\theta)$, is commonly referred to in the literature as the \textit{Held-Karp bound}.
This approach is typically incorporated into a branch-and-bound algorithm,
using this bound to  prune the search.
While computing a 1-tree is generally computationally efficient, the iterative adjustment of the $\theta$ multipliers can be computationally expensive. Our contribution is dedicated to mitigating this issue thanks to an unsupervised learning process.



\subparagraph{Example}

Figure~\ref{fig:HKrelax} illustrates the Held-Karp relaxation for a graph with an optimal
 TSP tour value of 62~(a). A 1-tree is computed on the original graph without Lagrangian multipliers, which yields a bound of 50~(b). Considering Equation~\eqref{eq:penalty} with $C = 2$, we obtain the following multipliers: ${\{\theta_1:0, \theta_2:4,\theta_3:0,\theta_4:-2,\theta_5:-2\}}$. 
The corresponding penalized 1-tree with Lagrangian multipliers modifying the edge costs provides a bound of 59, which is tighter~(c).

\section{Learning Held-Karp Lagrangian Multipliers} \label{sec:learnHK}

The Held-Karp bound $\mathsf{HK}(\theta)$ has two interesting properties: 
(1) it can be parameterized thanks to the $\theta$ Lagrangian multipliers,
and (2) it is always valid, meaning it will never exceed optimal TSP cost.
Both properties open the opportunity to use a learning-based approach to compute the bound. To do so, we propose to build a model ${\Phi_w: G(V,E) \to \mathbb{R}^{|V|}}$ able to directly predict all the $\theta$ multipliers 
for a TSP instance  given as input (i.e., a graph). The model
is parameterized with $p$ parameters $w = \{w_1,\dots,w_p\}$.
There are two benefits to this. First, it eliminates parts of the iterative process of \citet{heldkarp70} and saves execution time. Second, it allows us to potentially obtain tighter bounds. The process is illustrated in Figure~\ref{fig:HKGNNDiag}.

The goal is to find model parameters $w$ yielding the highest possible bound.
This corresponds to a maximization problem that can be solved by gradient-based optimization. The obtained bound is provably valid, regardless of the trained model's accuracy thanks to the second property.
We consider this a major strength of our contribution, 
as obtaining guarantees with machine learning in the context of combinatorial optimization is known to be a challenge~\cite{ijcai2021p610}.
We formulate the bound maximization problem and its gradient below.
\begin{equation}
\label{eq:obj}
\max_w \mathsf{HK}\big(\Phi_w(G)\big) \longmapsto \nabla_w \mathsf{HK}(\Phi_w(G)) 
\end{equation}
However, computing the gradient of this expression is not trivial,
as the bound is obtained by means of the 1-tree combinatorial structure $\mathcal{T}_\theta(c^\prime)$ (see Equation~\eqref{eq:bound}). As the tree is parameterized by $\theta$, the chain rule can
be applied to clarify the dependencies between
model parameters $w$ and Lagrangian multipliers $\theta$.
\begin{equation}
\label{eq:loss}
\nabla_w \mathsf{HK}\big(\Phi_w(G)\big) 
= \frac{\partial \mathsf{HK}(\Phi_w(G))}{\partial \theta} \times \frac{\partial \theta}{\partial w}
\end{equation}
The right term corresponds to the differentiation of the predictive neural network model~\cite{lecun2015deep} and is easily obtained by backpropagation~\cite{rumelhart1986learning}. On the other hand, the left term
requires to differentiate the expression depicted in Equation~\eqref{eq:bound} for all $\theta_i$ with $i \in V$. 
\begin{equation}
\label{eq:deriv}
 \frac{\partial \mathsf{HK}(.)}{\partial \theta} 
 = \frac{\partial \mathsf{cost}\big(\mathcal{T}_\theta(c^\prime)\big)}{\partial \theta} - 2 \frac{\partial \sum_{i = 1}^{^{|V|}} \theta_i}{\partial \theta}
\end{equation}
The
cost of the 1-tree (i.e., $\mathsf{cost}\big(\mathcal{T}_\theta(c^\prime)$)
corresponds to the weighted sum of the selected edges (i.e., variables $x_{i,j}$ for each $(i,j) \in E$). The cost $c^\prime_{i,j}$ defines the weights. 
\begin{equation}
\label{eq:deriv2}
 \frac{\partial \mathsf{HK}(.)}{\partial \theta} 
 = \frac{\partial \Big( \sum_{(i,j)}^{\in E} c^\prime_{i,j} x_{i,j} \Big)}{\partial \theta} - 2 \frac{\partial \sum_{i = 1}^{^{|V|}} \theta_i}{\partial \theta}
\end{equation}
Let us consider a specific multiplier $\theta_i$ associated to node $i \in V$ and let us unroll the cost as $c^\prime_{i,j} = c_{i,j} + \theta_i + \theta_j$ (see Equation~\eqref{eq:l7}). We can observe that the partial derivative of $\theta_i$ is non-zero only for the node itself and its adjacent edges, i.e. $(i,j) \in \delta(i)$.
\begin{align}
\label{eq:deriv3}
 \frac{\partial \mathsf{HK}(.)}{\partial \theta_i} 
 &= \frac{\partial  \sum_{(i,j)}^{\in \delta(i)} (c_{i,j} + \theta_i + \theta_j) x_{i,j} }{\partial \theta_i} - 2 \frac{\partial \theta_i }{\partial \theta_i}  \\
 &= \frac{\partial  \sum_{(i,j)}^{\in \delta(i)} \theta_i x_{i,j} }{\partial \theta_i} - 2 \frac{\partial \theta_i }{\partial \theta_i}   \\
 &= \mathlarger{\sum}_{(i,j) \in \delta(i)} x_{i,j} - 2 
\end{align}
This gives the partial derivative for each  $\theta_i$ and 
allows us to maximize a bound obtained by a neural network
directly with gradient ascent. Interestingly, this signal 
is non-zero when the degree of the node is different than 2 in the 1-tree.
This is aligned with the intuition that we want to adjust the multipliers
of conflicting nodes.

The training procedure is formalized in
Algorithm~\ref{algo:trainHK}. It gives as output the parameters $w$
of the trained neural network $\Phi_w$. We note that this training loop
can be easily improved with standard techniques in deep learning, such as mini-batches or using another gradient-based optimizer, such as Adam~\cite{kingma2014adam}. 
Unlike gradient descent, we aim to maximize the bound, explaining the $+$ term at Line 10. We highlight that the training is \textit{unsupervised}
as it does not require ground truth on known tight bounds for training the model, nor the corresponding Lagrangian multipliers.
Finally, two aspects of the methodology require clarification: the architecture of the neural network $\Phi_w$ and the construction of the training set $\mathcal{D}$.
Both are discussed in the following sections.
\begin{algorithm}[!ht]
\begin{algorithmic}[1]

\STATE $\triangleright$ \textbf{Pre:} $D$ is the set of instances used for training.
            
\STATE $\triangleright$ \textbf{Pre:} $\Phi_w$ is the differentiable model to train. 

\STATE $\triangleright$ \textbf{Pre:} $w$ are randomly initialized parameters. 

\STATE $\triangleright$ \textbf{Pre:} $K$ is the number of training epochs.

~

\FOR{$k \mathbf{~from~} 1 \mathbf{~to~} K$}

\STATE $G := \mathsf{SampleFromTrainingSet}(\mathcal{D})$
\STATE $\theta := \Phi_w(G)$

\STATE $\mathcal{T}_\theta(c^\prime) :=  \mathsf{HeldKarpRelaxation}(G, \theta)$

\STATE $\mathsf{HK}\big(\theta \big) := \mathsf{cost}\big(\mathcal{T}_\theta(c^\prime)\big) - 2\sum_{i = 1}^{|V|} \theta_i$

\STATE $w := w + \nabla_w \mathsf{HK}\big(\Phi_w(G)\big) $

\ENDFOR

\RETURN{$w$}

\end{algorithmic}
\caption{Training step from an input graph $G(V,E)$.}
\label{algo:trainHK}
\end{algorithm}

\subsection{Training Set Construction}
The training is carried out from a dataset $\mathcal{D}$ consisting of a set of graphs $G(V,E)$ serving as TSP instances. The graphs can either be obtained from historical problem instances (e.g., previous routing networks and costs for a delivery company) or randomly generated. Each graph has six features $f_i$ for each node $i \in V$ and three features $k_{i,j}$ for each edge $(i,j) \in E$. The features are summarized in the supplementary material.
Although most of the features are relatively standard, $k^2_{i,j}$ and $k^3_{i,j}$ introduce the notion of \textit{mandatory} and \textit{forbidden} edges. In the context of a branch-and-bound algorithm,
some decision variables are fixed after branching operations. An edge is mandatory if it 
must be part of the TSP solution (i.e., $x_{i,j} = 1$) and it is forbidden if it cannot be in the solution (i.e., $x_{i,j} = 0$). 
This information is crucial as we plan to compute bounds several times during
a branch-and-bound execution, with the motivation to leverage partial solutions
to get better bounds. 

A direct observation is that there are no fixed edges at the root node of a branch-and-bound tree, and consequently, for none of the instances in the training set. This causes a distributional shift between instances
used for the training (only at the root node)
and the ones occurring at the testing phase (also inside the branch-and-bound tree). To address this limitation, we propose to enrich the training set with 
partially solved TSP instances extracted from explored branch-and-bound nodes. 
In practice, it is done by fixing a threshold  $k \in \mathbb{N}^+$ 
on the number of nodes to consider in the training set. This makes
the computation tractable as it avoids considering all the nodes of an exponentially sized tree search.

\begin{table*}[!ht]
\resizebox{\textwidth}{!} 
{ 
\centering
\renewcommand{\arraystretch}{1.3}
\begin{tabular}{l rrrrr rrrrr }\toprule

\multirow{2}{*}{Configuration} & \multicolumn{5}{c}{Branch-and-bound with standard Held-Karp ($\mathsf{HK}$)} & \multicolumn{5}{c}{Branch-and-bound with our approach ($\mathsf{GNN} + \mathsf{HK}$)}  \\
 \cmidrule(lr){2-6} \cmidrule(lr){7-11}
 & Time  (sec.) & \# solved (/50) & PDI & Filt. (\%) & Opt. gap (\%) &  Time  (sec.)  & \# solved (/50)  & PDI & Filt. (\%) & Opt. gap (\%)  \\

\midrule
\textsf{Random100} & 559 & 41/50 & 1127k & 75.9 & 0.88 & 497 (- 11\%)
& 46/50 (+ 10\%) & 965k (- 14\%) & 77.7 (+ 2\%) & 0.48 (- 45\%) \\
\textsf{Random200} & 1800 & 0/50 & 4.71m & 67.8 & 1.82 & 1800 & 0/50 (+ 0\%) & 4.26m (- 10\%) & 70.6 (+ 4\%) & 0.59 (- 68\%) \\
\textsf{Clustered100} & 643 & 38/50 & 497k & 17.7 & 0.19 & 590 (- 8\%) & 40/50 (+ 5\%) & 470k (- 5\%) & 20.3 (+ 15\%) & 0.08 (- 58\%) \\
\textsf{Clustered200} & 1800 & 0/50 & 922k & 9.9 & 0.68 & 1800 & 0/50 (+ 0\%) & 690k (- 25\%) & 12.6 (+27\%) & 0.38 (- 44\%) \\
\textsf{Hard} & 1800 & 0/6 & 9.59M & 6.4 & 0.32 & 1800 & 0/6 (+ 0\%)& 9.36M (- 2\%) & 6.5 (+1\%) & 0.31 (- 3\%) \\
\bottomrule
\end{tabular}
}
\caption{Comparison of our approach ($\mathsf{GNN+HK}$) with the standard branch-and-bound of \citet{benchimol2012improved} ($\mathsf{HK}$).
The primal bound is 2\% above the optimal solution cost computed with Concorde~\cite{Applegate06}.
The statistics considered are: the execution time up to a timeout of 1,800 seconds (\textit{Time}), the number of
instances solved to optimality with proof (\textit{\# solved}), the primal-dual integral (\textit{PDI}), the percentage of edges filtered (\textit{Filt.}) and the optimality gap for unsolved instances (\textit{Opt. gap}). The relative improvement compared to the baseline is also depicted.}
\label{tab:main-results}

\end{table*}

\subsection{Graph Neural Network Architecture}
\label{sec:genFact}

A TSP instance exhibits a natural graph structure.
For this reason, we built the model $\Phi_w$ with a \textit{graph neural network}~\cite{scarselli2008graph,kipf2016semi} (GNN).
This architecture has been widely in related works for the TSP,
thanks to their ability to handle instances of different size, to leverage
node and edge features, etc. In its standard version, 
GNNs are dedicated to computing a vector 
representation of each node of the graph. 
Such a representation is commonly referred to as an \textit{embedding}.
The embedding of a specific node is computed by iteratively transforming and aggregating information from the neighboring nodes.
Each aggregation operation is referred to as a layer of the GNN
and involves weights that must be learned. This operation
can be performed in many ways, and there exist in the 
literature different variants of GNNs. 
An analysis on the performances of various architectures is
proposed by~\citet{dwivedi2023benchmarking}. 
Our model is based on the \textit{edge-featured graph attention network}~
\cite{Wang21egat} which is a variant of 
the well-known \textit{graph attention network}~\cite{velickovic2018graph} dedicated 
to handle features on the edges. The whole architecture 
is differentiable and can be trained with backpropagation.

Let $G(V,E)$ be the input graph, $f_i \in \mathbb{R}^6$
 be a vector concatenating the 6 features of a node $i \in V$, 
and $k_{i,j} \in \mathbb{R}^3$ be a vector concatenating the three features of an edge $(i,j) \in E$. 
The GNN architecture is composed of a set of layers $L$.
Let $h^l_i \in \mathbb{R}^d$ 
be a $d$-dimensional vector representation of a node $i \in V$ at layer 
$l \in L$, and let $h^{l+1}_i \in \mathbb{R}^{d^\prime}$ 
a $d^\prime$-dimensional vector representation of $i$ at the next layer.
The inference process consists in computing the next representation
($h^{l+1}_i$) from the previous one $h^{l}_i$ for each node $i$.
The initial representation is initialized with the 
initial features of the node, i.e. $h^{1}_i = f_i$ for each $i \in V$.
The computation is formalized in Equations~\eqref{eq:gnn1} to~\eqref{eq:gnn3},
where $w_1^l$ and $w_2^l$ are two weight tensors that need to be trained for each layer $l \in L$.
Equation~\eqref{eq:gnn1} shows the message passing operation in a layer. 
Each node $i$ aggregates information of all its neighbors $\mathcal{N}(i)$.
The aggregation is subject to parameterized weights $w^l_1$ 
and a \textit{self-attention score}
$\alpha^l_{i,j}$. This score allows the model to put different weights on the incoming messages from neighboring nodes.
We note that the attention integrates information about the node itself ($h^l_i$), its neighbor ($h^l_j$), and
the features attached to the adjacent edge ($k_{i,j}$). Such information is concatenated ($.\|.$) into a single vector. Non-linearities are added after each aggregation and the final node embeddings 
$h^{|L|}_i$ are given as input to a  fully-connected neural network ($\mathsf{FCNN}$) outputing
the corresponding $\theta_i$ multiplier for each $i \in V$.
The GNN has 3 graph attention  layers with a hidden size of 32 and the fully-connected neural network has
2 layers with 32 neurons.
\begin{equation}
\label{eq:gnn1}
h^{l+1}_i = \mathsf{ReLU}\Big( \sum_{j \in \mathcal{N}(i)} \alpha^{l}_{(i,j)} w^l_1 h^{l}_j  \Big) ~ ~ \forall i \in V ~ \land ~  \forall l \in L
\end{equation}
\begin{equation}
\label{eq:gnn2}
\alpha^{l}_{(i,j)} = \mathsf{Softmax}\bigg( \mathsf{LeakyReLU} \Big( w_2^{l} \times 
\big(h^l_i \big\|  k_{i,j}  \big\| h^{l}_j  \big) \Big) \bigg)
\end{equation}
\begin{equation}
\label{eq:gnn3}
\theta_v = \mathsf{FCNN}\big(h^{|L|}_i\big)  \hspace{2cm} \forall i \in V
\end{equation}

\begin{figure*}[ht!]
  \centering

\begin{subfigure}[b]{0.25\textwidth}
\includegraphics[width=0.96\textwidth]{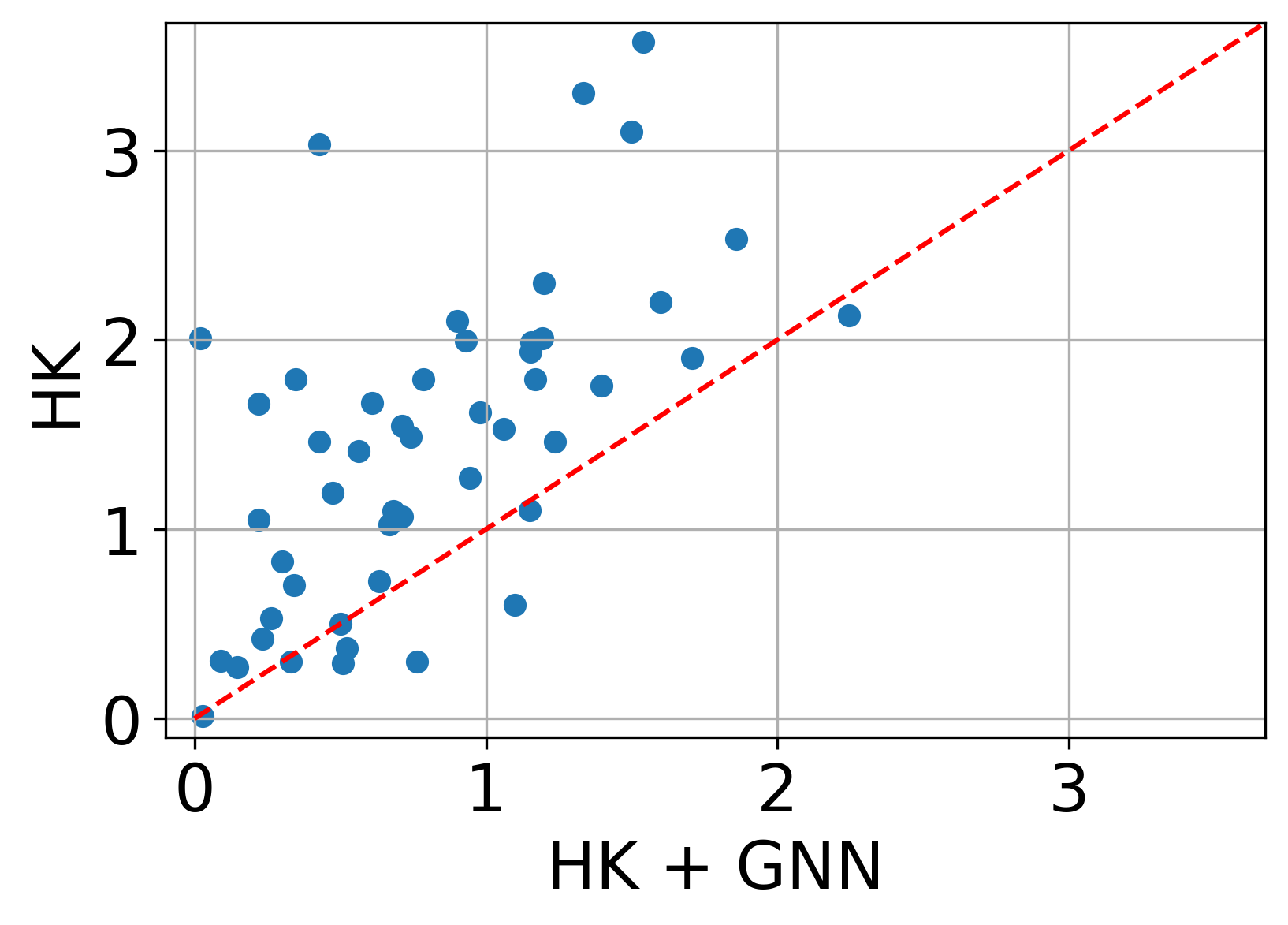}
\caption{Results on $\mathsf{Random200}$.}
\label{subfig:scat-1}
\end{subfigure}
\hfill
\begin{subfigure}[b]{0.25\textwidth}
\includegraphics[width=\textwidth]{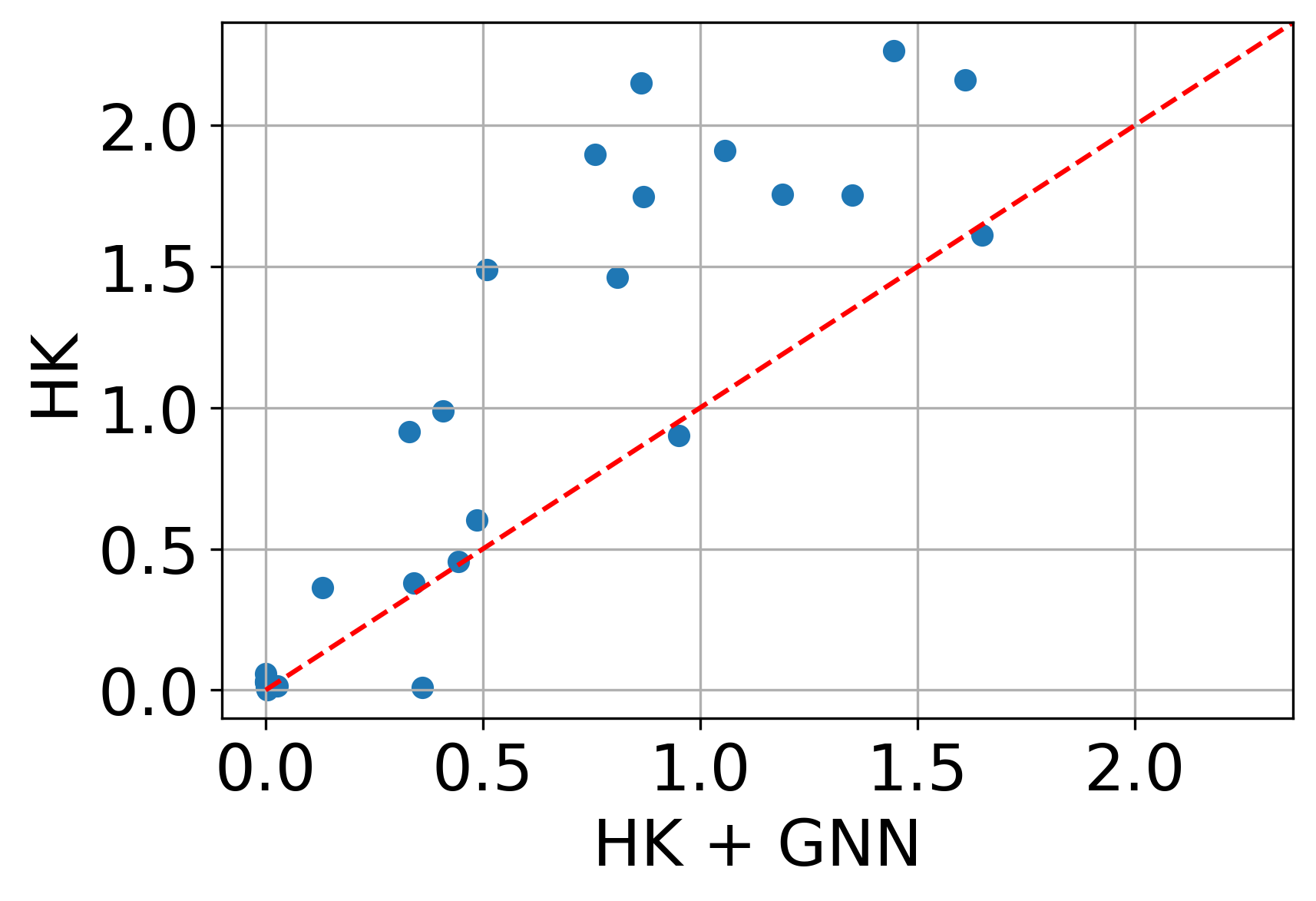}
\caption{Results on $\mathsf{Clustered200}$.}
\label{subfig:scat-2}
\end{subfigure}
\hfill
\begin{subfigure}[b]{0.25\textwidth}
\includegraphics[width=\textwidth]{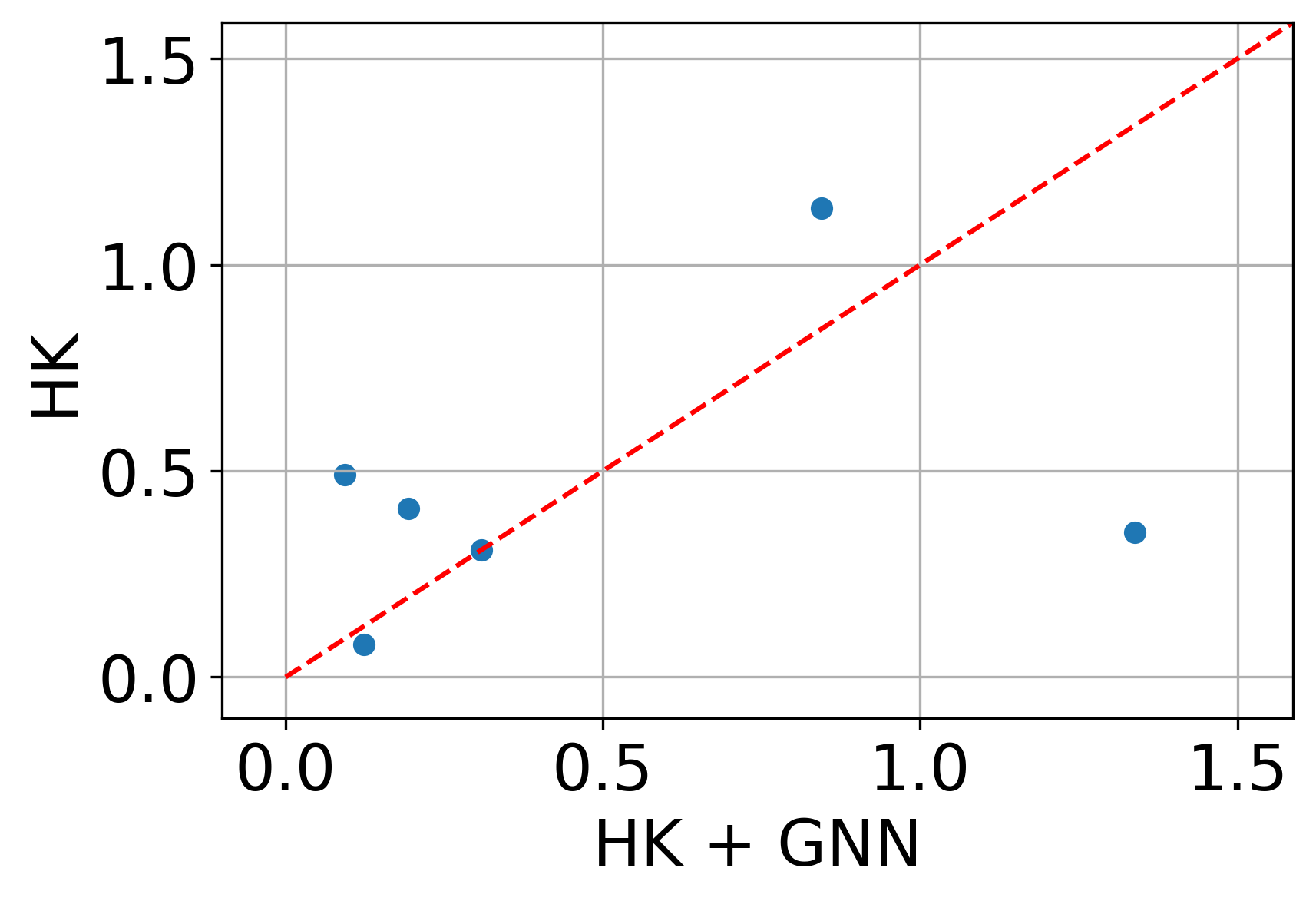}
\caption{Results on $\mathsf{Hard}$.}
\label{subfig:scat-3}
\end{subfigure}

  \caption{Scatter plots comparing the optimality gap (\%) for $\mathsf{HK}$ and $\mathsf{HK+GNN}$ on the three hardest configurations.} \label{fig:scatterplot}
\end{figure*}

\section{Experimental Evaluation} \label{sec:appHK}
The goal of the experiments is to evaluate the efficiency of
the approach to speed-up a TSP solver based on branch-and-bound and constraint programming~\cite{benchimol2012improved}.
To do so, the learned bounds are integrated into the Held-Karp relaxation used by the weighted circuit constraint~\cite{beldiceanu1994introducing}
and are used to filter unpromising edges. The model is used only for the $10$ first levels of the branch-and-bound tree (parameter $k$). We refer to $\mathsf{HK}$ for the standard solver of \citet{benchimol2012improved} and to
$\mathsf{HK+GNN}$ for the one we introduce.

\begin{table*}[!ht]
\resizebox{\textwidth}{!} 
{ 
\centering
\renewcommand{\arraystretch}{1.3}
\begin{tabular}{l rrrrr rrrrr }\toprule

\multirow{2}{*}{Configuration} & \multicolumn{5}{c}{Branch-and-bound with standard Held-Karp ($\mathsf{HK}$)} & \multicolumn{5}{c}{Branch-and-bound with our approach ($\mathsf{GNN} + \mathsf{HK}$)}  \\
 \cmidrule(lr){2-6} \cmidrule(lr){7-11}
 & Time  (sec.) & \# solved (/50) & PDI & Filt. (\%) & Opt. gap (\%) &  Time  (sec.)  & \# solved (/50)  & PDI & Filt. (\%) & Opt. gap (\%)  \\

\midrule
\textsf{Random100} & 209 & 48/50  & 591k & 92.1 & 0.05 & 203 (- 3\%) & 48/50 (+ 0\%) & 544k (- 7\%) & 92.2 (+ 0\%) & 0.05 (- 0\%) \\
\textsf{Random200} & 1800 & 0/50  & 4.44m & 85.6 & 1.16 & 1800 & 0/50 (+ 0\%) & 4.11m (- 7\%) & 89.1 (+ 4\%) & 0.43 (- 62\%) \\
\textsf{Clustered100} & 112 & 48/50  & 103k & 25.4 & 0 & 110 (- 1\%) & 49/50 (+ 2\%) & 100k (- 3\%) & 25.5 (+ 0\%) & 0 (- 0\%) \\
\textsf{Clustered200} & 1800 & 0/50  & 713k & 14.5 & 0.43 & 1800 & 0/50 (+ 0\%) & 644k (-9\%) & 16.8 (+15\%) & 0.30 (- 30\%) \\
\textsf{Hard} & 1800 & 0/6 & 7.59M & 17.8 & 0.26 & 1800 & 0/6 (+ 0\%) & 7.02M (- 7\%) & 18.0 (+ 1\%) & 0.19 (- 24\%) \\
\bottomrule
\end{tabular}
}
\caption{Comparison of our approach ($\mathsf{GNN+HK}$) with the standard branch-and-bound of \citet{benchimol2012improved} ($\mathsf{HK}$) for optimality proof.
Compared to Table~\ref{tab:main-results}, the primal bounds correspond now the cost of the optimal solution.}
\label{tab:opt-results}

\end{table*}

\subsection{Experimental Protocol}

\subsubsection{Datasets}
Five datasets of different complexity are considered. They correspond to variants of the metric TSP (i.e., the graphs are complete and the distances are euclidean) on which the cities are localized with different patterns.

\begin{enumerate}
    \item \textsf{Random100} (and \textsf{200}): the cities (100 or 200) are  uniformly generated  in the  $[0,1]^2$ plan. 
    \item \textsf{Clustered100} (and \textsf{200}): inspired by~\citet{fischetti1989additive}, five clusters are uniformly generated  in the  $[0,1]^2$ plan. Then, the cities  (100 or 200) are uniformly generated  inside the 0.1-radius circles around each cluster.
    \item \textsf{Hard}: introduced by~\citet{hougardy2021hard}, these 50 instances ranging from 52 to 199 cities have been generated to have a large integrality gap and are provably hard to solve for branch-and-bound methods.
\end{enumerate}

A test set of 50 instances is built for each configuration and is used for evaluation.
For the last dataset, as it is relatively small, only 6 instances are taken for  evaluation.

\subsubsection{Training}
The training phase corresponds to the execution of Algorithm~\ref{algo:trainHK}. 
A specific model is trained for the five configurations. The training sets for
 \textsf{Random} and \textsf{Clustered} have 100 instances sampled from the given generating scheme.
 For the \textsf{Hard} dataset, 40 instances uniformly sampled from the 50 available instances are taken.
 For each instance, 10 subgraphs are generated (parameter $k$) and are added to the training set. They correspond to partially solved instances that could be found inside the branch-and-bound tree.
We highlight that we do not need to label the training instances with known multipliers as the learning is unsupervised.
Training time is limited to 4 hours on a AMD Rome 7502 2.50GHz processor with 64GB RAM. No GPU has been used.
Models are trained with a single run and we observed the convergence
of the loss function on a validation set of 20 instances.
 The Adam optimizer~\cite{kingma2014adam} with a learning rate of $10^{-3}$ has been used.

\subsubsection{Implementation}
The graph neural network has been implemented with \textit{deep graph library}~\cite{wang2019dgl} and \textit{Pytorch}~\cite{paszke2019pytorch}. During
the training, the minimum spanning trees  have been computed with \textit{NetworkX}~\cite{hagberg2008exploring}.
We use the \textit{C++} implementation of \citet{benchimol2012improved} for the branch-and-bound solver. The interface between the \textit{python} and the \textit{C++} code has been done with native functions from both languages ; the implementation and the datasets used are released with the
permissive MIT open-source license\footnote{https://github.com/corail-research/learning-hk-bound}.

\subsubsection{Hyperparameters}
The branch-and-bound  has been configured with the standard settings and most of the hyperparameters used 
follow the recommended values. No hyperparameter tuning  has been carried out due to our limited resources.
A notable exception is the threshold $k$ on the maximum number of nodes  which has an important impact on the performances.
We tested the values $k \in \{1, 5, 10, 20, 50\}$ and  selected $k=10$ as it provided the best results and trade-off between accuracy and computation time.

\subsubsection{Evaluation Metrics}
Five metrics are reported in the results: 
(1) the execution time to prove the optimality of a solution,
(2) the number of instances solved to optimality, 
(3) the percentage of edges filtered before branching through the propagation,
(4) the optimality gap for unsolved instances, and (5) the primal-dual integral.
This last metric measures the convergence of the dual bound and the primal bound over the whole solving time~\cite{berthold2013measuring}. 
Each instance is solved only once per experiment as no randomness is involved in the execution.

\subsection{Empirical Results}

\subsubsection{Main Results: Quality of the Learned Bounds}
Table~\ref{tab:main-results} summarizes the results for \textsf{HK} and \textsf{HK+GNN} on the five datasets. Values
are averaged for each configuration. First, we can observe that our approach gives consistently better results
on all the metrics compared to the baseline. As expected, it provides better filtering on the edges.
this is reflected by a higher number of solved instances, a reduced execution time, and a reduced optimality gap for unsolved instances.
The primal-dual integral confirms that tighter dual bounds are obtained during the search.
Second, we notice that the improvements on the \textsf{Hard} dataset are more modest. 
This can be explained by the fact that they are designed to be challenging. It is consequently more difficult to get improvements on these ones.

\subsubsection{Analysis: Focus on Individual Instances}
Figure~\ref{fig:scatterplot} provides an analysis of the optimality gap for the three hardest configurations by means of scatter plots.
Each dot corresponds to a specific instance. When a dot is upper than the diagonal, it means that our approach provided better results 
than the baseline. Unlike the previous experiments, it provides insights about the robustness of the method.
For the majority of the instances, our approach gave better or similar results, except for one instance in the \textsf{Hard} dataset.

\subsubsection{Analysis: Addressing the Optimality Proof}
This next experiment evaluates the ability of proving the optimality of a solution \textit{only once this solution has been found}. 
Concretely, instead of taking a reasonable upper bound of 2\%, 
we assume that the optimal solution has been found and we use it as a perfect upper bound. The task is to close the search by proving the optimality of the solution. Results are summarized in Table~\ref{tab:opt-results}. 
In such a situation, the improvements with the baseline are still 
positive, especially for the largest and hardest configurations.
This shows the potential of learning dual bounds to accelerate optimality proofs.

\subsubsection{Analysis: Generalization Ability}
This last experiment analyzes how the models are able to generalize to new configurations, either for a higher number of cities or with another generation scheme.
Concretely, we considered four configurations (\textsf{Random100}, \textsf{Random200}, \textsf{Clustered100} and \textsf{AllDataset}, the latter being trained on the instance of all datasets) and evaluated them on \textsf{Clustered200}.
Results are presented in Table~\ref{tab:generalizationClustered200}. 
Although the performance of the specific model is not reached, we
observe that the models trained on the other distributions are still able to outperform the standard model.
Training a model on all datasets (\textsf{AllDataset}) allows to improve upon out-of-distribution models but does not achieve the performance of the specialized model. This confirms the intuitive benefit to know beforehand the distribution of the instances to solved.

\begin{table}[!ht]
\centering
\resizebox{\columnwidth}{!} 
{
\begin{tabular}{l rrr }\toprule

\multirow{2}{*}{Model} & \multicolumn{3}{c}{Branch-and-bound with $\mathsf{GNN} + \mathsf{HK}$}  \\
 \cmidrule(lr){2-4}
& PDI & Filt. (\%) & Opt. gap (\%)  \\

\midrule
\textsf{Clustered200} & 690k & 12.6 & 0.38 \\
\midrule
$\mathsf{HK}$ without $\mathsf{GNN}$ & 922k & 9.9  & 0.68 \\
\midrule
\textsf{Clustered100}& 817k & 11.1 & 0.54 \\
\textsf{Random100}   & 845k & 10.4 & 0.61 \\
\textsf{Random200}   & 784k & 11.3 & 0.49 \\
\textsf{AllDataset}  & 722k & 12.0 & 0.45\\
\bottomrule
\end{tabular}
}
\caption{Analysis of the generalization. The different  models are used to solve \textsf{Clustered200} graphs.}
\label{tab:generalizationClustered200}
\end{table}

\section{Conclusion}

Learning-based methods have been extensively considered to provide approximate solutions to combinatorial optimization problems, such as the travelling salesman 
However, learning to obtain dual bounds has been less considered in the literature and is much more challenging as there is no trivial way to ensure that the bounds obtained are valid. 
This paper introduces an unsupervised learning approach, based on graph neural networks and the Held-Karp Lagrangian relaxation, to tackle this challenge. The core idea is to predict accurate Lagrangian multipliers and employ them as a warm start for generating Held-Karp relaxation bounds. These bounds are subsequently used to enhance the filtering level of the weighted circuit global constraint and improve the performances of a branch-and-bound algorithm.
The empirical results on different configurations of the TSP showed that the learning component can significantly improve the algorithm.
Although centered on the TSP, we note that weighted circuit global constraint could be used for other, and more challenging, 
TSP variants including time windows or time-dependent costs. Analyzing how these variants can be handled efficiently is part of our future work.

\bibliography{arxiv-submission}

\newpage

\section{Appendix A: Summary of the Features}

The input graphs are enriched with features representing the problem, the state of the branch-and-bound search. The goal is to provide a rich representation of the nodes and edges for accurate Lagrangian multipliers generation. The features we used are presented in Table~\ref{tab:features}.

\begin{table}[!ht]

\centering
\resizebox{\textwidth}{!} 
{ 
\renewcommand{\arraystretch}{1.3}
\begin{tabular}{c|lll}\toprule
& Symbol & Formalization & Description \\ 
\midrule
\midrule
\multirow{5}{*}{\shortstack{Features on nodes $i$ \\ for each $i \in V$}} &  $f^1_i,f^2_i \in \mathbb{R}^2 $   & $\mathsf{xPos}(i),\mathsf{yPos}(i)$ & 2-dimensional coordinate of the node. \\
& $f^3_i \in \mathbb{R} $   & $\frac{1}{|V|} \sum\nolimits_{j=1}^{|V|}\| \mathsf{coord}(i) - \mathsf{coord}(j) \|_2$ & Average euclidean distance with the other nodes.\\
&  $f^4_i \in \mathbb{R}$   & $\min\nolimits_{j \neq i} ~ (f^3_1,\dots,f^3_j,\dots,f^3_{|V|} )$ & Distance to the nearest node in the graph. \\
& $f^5_i \in \mathbb{N}^+$   & $\mathsf{deg}(i)$ & Degree in terms of incoming and outgoing edges. \\
&  $f^6_i \in \{0,1\}$   & $1 \mathsf{~iff~} i = 1, 0 \mathsf{~otherwise}$ & Binary value indicating if it is the excluded node in $G \backslash \{1\}$. \\
\midrule
\midrule
 \multirow{3}{*}{\shortstack{Features  on edges $(i,j)$ \\ for each $(i,j) \in E$}}  & $k^1_{i,j} \in \mathbb{R}$   & $c_{i,j}$ & The cost of the edge. \\
& $k^2_{i,j} \in \{0,1\}$   & $1 \mathsf{~iff~} (i,j) \mathsf{~is~forbidden}, 0 \mathsf{~otherwise}$ & Binary value indicating if the edge is \textit{forbidden}. \\
& $k^3_{i,j} \in \{0,1\}$   & $1 \mathsf{~iff~} (i,j) \mathsf{~is~mandarory}, 0 \mathsf{~otherwise}$ & Binary value indicating if the edge is \textit{mandatory}.\\
\bottomrule
\end{tabular}
}
\caption{Features used in an input graph $G(V,E)$.}
\label{tab:features}
\end{table}

\end{document}